\documentclass[11pt,a4paper]{article}
\usepackage[hyperref]{acl2019}
\usepackage{times}
\usepackage{latexsym}
\usepackage{graphicx}
\usepackage{url}
\usepackage{multirow}
\usepackage{pbox}
\usepackage{todonotes}

\usepackage{url}

\aclfinalcopy 

\title{ Supervised Contextual Embeddings for Transfer \\ Learning in Natural Language Processing Tasks}

\author{Mihir Kale\thanks{equal contribution}, Aditya Siddhant\footnotemark[1], Sreyashi Nag\footnotemark[1], Radhika Parik\footnotemark[1], \\
        \bf{Matthias Grabmair, Anthony Tomasic} \\
  \{mihirsak, asiddhan, sreyashn, rparik, mgrabmai, tomasic\} @cs.cmu.edu \\
  Language Technologies Institute, Carnegie Mellon University}

\date{}

\begin{document}
\maketitle
\begin{abstract}
  Pre-trained word embeddings are the primary method for transfer learning in several Natural Language Processing (NLP) tasks. Recent works have focused on using unsupervised techniques such as language modeling to obtain these embeddings. In contrast, this work focuses on extracting representations from multiple pre-trained supervised models, which enriches word embeddings with task and domain specific knowledge. Experiments performed in cross-task, cross-domain and cross-lingual settings indicate that such supervised embeddings are helpful, especially in the low-resource setting, but the extent of gains is dependent on the nature of the task and domain. We make our code publicly available. \footnote{https://github.com/asiddhant/taskonomy-nlp}
\end{abstract}

\section{Introduction}
Named entity recognition, semantic role labelling, relation extraction etc. can be thought of as primary tasks necessary for solving high level tasks like question answering, summarization etc. However, labelling large amounts of data at this granularity is not only prohibitively expensive, but also unscalable. Given that high performance models for these tasks already exist, it is desirable to leverage them for other language understanding tasks.
 
Next, consider the domain adaptation setting where some domains have a lot of data, while others do not. A model for a low-resource domain would benefit from information in \textit{expert} models trained on other data rich domains. Finally, consider the setting of cross-lingual adaptation, a common problem for personal assistants expanding to more languages. As the number of languages increases, it becomes unfeasible to obtain human annotated data. Again, the need to adapt to low-resource languages can be met by leveraging models that already exist for high-resource languages.

Motivated by the above scenarios, we propose a simple method to transfer (1) \textit{supervised knowledge}, from (2) \textit{multiple sources}, (3) in an \textit{easy to implement} manner. In our approach, this knowledge is extracted from source models in the form of contextual word embeddings. We treat preexisting models as embedding extractors, which are used to extract token level representations for an input sentence. These representations are then combined via a task specific convex combination.

Unsupervised transfer learning methods such as ELMo have shown great success for a variety of tasks \citep{peters2018deep}. While they have the advantage of being trained on very large corpora, the training objectives are \textit{unsupervised}. We show that in low-resource settings especially, leveraging representations from \textit{multiple pre-trained supervised} models in related tasks, domains or languages can prove to be beneficial.

The common way of supervised transfer learning via fine-tuning can transfer information only from a single source task \citep{mou2016transferable}. One way to incorporate information from multiple external sources is via multi-task learning \citep{hashimoto2017joint,ruder2017overview}. The limitations of multi-task learning are the need for labelled data for the source models, longer training times and complex design decisions (weighing the losses for each task, sampling strategies, and choice of architecture). In contrast, our plug-and-play approach is simple and does not assume availability of source model data at training time. Finally, our approach also provides some interpretability (through the parameters of the convex combination) into which source tasks or domains are important for which other tasks and domains.

\section{Related Work}
Our work aligns most with the following three directions of research.
\paragraph{Unsupervised transfer learning} Embeddings such as GloVe and FastText have become an integral part of the modern NLP pipeline \citep{pennington2014glove,bojanowski2016enriching}. Over the last year, language model based deep contextualized embedding methods such as ELMo have shown substantial improvements over their shallow counterparts, heralding a new era of word representations \citep{peters2018deep}.
\paragraph{Supervised transfer learning} CoVe \citep{mccann2017learned} and InferSent \citep{conneau2017supervised} extract embeddings from encoders pre-trained for Machine Translation and Natural Language Inference respectively.
\citet{mihaylov2017neural} transfer low-level skills such as textual entailment, NER, paraphrase detection and question type classification into a reading comprehension model. 
\paragraph{Multi-source transfer learning} In terms of modelling approach, our work is similar to \citet{kim2017domain} , where the authors use multiple existing models for domain adaptation for spoken language understanding. In comparison, our work focuses not just on the domain adaptation, but also the cross-task and cross-lingual settings. In another work, \citet{coates2018frustratingly} create meta-embeddings from multiple embeddings like GloVe, Fasttext etc.

\section{Approach} \label{approach}
Most deep learning models can be thought of as having an encoder $E$ and decoder $D$. For example in a Deep-SRL model \citep{He2017DeepSR}, stacked bidirectional LSTM constitutes $E$, while $D$ is the softmax layer. Assume $K$ existing supervised models either for different tasks or different domains $M_1,...,M_K$ and corresponding encoders $E_1,...,E_K$. Given a sentence of N tokens ($t_1,t_2,...,t_N$), we feed these tokens to the $K$ different encoders and get $K$ different representations for each token. We denote the encoder output of the $k$th model for the $n$th token by $h_n^{k}$. Each encoder generates representations specialized for the task, domain, or language it was trained for. Since our approach assumes no explicit information about the encoders of the model, they can be of varying dimensions and use different underlying architectures. Evidently, they would also be in different vector spaces and therefore we first use a projection layer to bring all of them in the same vector space. The parameters of these projection layers $W_1,...W_K$ are learned along with the target model parameters. $W_k$ projects $h_n^k$ to a fixed $D$ dimensional vector $g_n^k$.

For inclusion in a downstream model, we aggregate the projection layer output of all the different source models into one vector. Several aggregation schemes can be employed : pooling, convex combination, attention etc. We choose the simple yet interpretable convex combination approach, as described below.

\textbf{Convex Combination}: This technique is similar to one used by ELMo \citep{peters2018deep}. We use a softmax normalized weight $s_k$ corresponding to each of the different representations of the word, add them up and use a scalar parameter $\gamma$ that scales up the whole vector. The embedding $O_n$ for the $n$th word comes out to be:
    $$ O_n = \gamma ~ \sum_{k=1}^K s_k ~ g_n^k $$
    This approach adds $K+1$ trainable parameters to the model. An advantage of combining the representations in this manner is that the size of the embedding is fixed irrespective of the number of source models used.

Once we get a combined representation, it can be used in the target model just like any other embedding. In our experiments, we concatenate these embeddings with traditional GloVe or ELMo embeddings.

\section{Experimental Setup}
We use the proposed supervised contextual embeddings along with GloVe and ELMo embeddings in three knowledge transfer settings. 

\paragraph{Cross-task transfer} In this setting, we transfer knowledge to a target task from models trained on multiple source tasks. We transfer into Semantic Role Labeling (SRL) task using  Constituency Parsing (CP), Dependency Parsing (DP) and Named Entity Recognition (NER) as source tasks. The choice of SRL as a target task, with source embeddings from CP, DP and NER models,  is inspired by the popular use of explicit syntactic parsing features for SRL. We use OntoNotes 5.0 \citep{pradhan2012conll} dataset to train the SRL target tasks. We use the stacked alternating LSTM architechture for SRL as per \citet{He2017DeepSR}. On the source side, the DP model is based on \citet{dozat2016deep} and CP on \citet{Stern2017AMS}. For most of the source models, we use off-the-shelf,  pre-trained models provided by AllenNLP \footnote{https://allennlp.org/models}. We refer readers to \citet{peters2018deep} for further description of model architectures for the various tasks.

\paragraph{Cross-domain transfer} Here, we study the applicability of our method in the cross-domain setting. The target task is same as the source tasks, but instead, the domains of the source and target models are different. For this set of experiments, our task is NER and we use the OntoNotes 5.0 dataset which comes with annotations for multiple domains. Though NER is an easier task, we chose it as the target task for the cross-domain setting as even state of the art NER models may perform poorly for a  data-scarce domain. We choose the target domain as web blogs and the source domains are newswire, broadcast conversation, telephone conversation, magazines and broadcast news. Note that the samples in the validation and test sets are also limited to the web blogs domain only. We use an LSTM-CRF architechture with 1 LSTM layer for NER as per \citet{peters2017semi}.

\paragraph{Cross-lingual transfer} From the CoNLL shared tasks, we obtain NER datasets for English, Spanish, German and Dutch \citep{tjong2003introduction}. We consider two scenarios with German and Spanish as the target languages and the remaining 3 as source languages. To facilitate the input of sentences into models from other languages with different scripts, we rely on cross-lingual embeddings provided by MUSE \citep{conneau2017word}. The NER model architecture is the same as the one used for the cross-domain experiments.

To study the effectiveness of our approach in the \textbf{low resource setting}, in addition to the full datasets, we also run experiments on smaller training subsets. Similar to \citet{mulcaire2018polyglot}, we create random subsets of 1,000 and 5,000 samples to simulate a low resource setting. In all the aforementoiend settings, the source task models are trained on their complete datasets.

\paragraph{Hyperparameters} We use the Adam optimizer (lr=0.001) for all our experiments. We run our target models for 50 epochs in SRL tasks and 75 epochs for NER tasks. Batch size is kept at 8 for the 1k data setting and 16 for 5k data setting. The dimensions of the GloVe and ELMo embeddings are 100 and 1024 respectively. The output dimension of the projection layer in all settings for supervised embeddings is 300.

\section{Results and Discussion}
Cross-task SRL results (with GloVe and ELMo in  1k, 5k and full data settings) have been tabulated in Table \ref{tab:srl_cross_task}. Table \ref{tab:ner_cross_domain} has the results for cross-domain NER and Table \ref{tab:cross_ling} shows the results for cross-lingual transfer on NER. All the reported numbers are F1 scores.

\paragraph{Cross-task SRL} With GloVe embeddings, adding the supervised embeddings gives us significant improvements in F1 scores $\sim$ 5\% for 1k and $\sim$ 7\% for 5k examples. When we use the entire dataset, adding supervised embeddings provides no performance gains. Examining the learned source task weights in the 1k setting, we find that weights for CP, DP and NER have values 0.41, 0.41 and 0.18 respectively which shows that SRL benefits greatly from syntactic tasks like CP and DP. This is in agreement with state-of-the-art SRL models \citep{strubell2018linguistically, marcheggiani2017encoding} which rely on syntactic features.

When we replace GloVe with ELMo representations, we see that the baseline model improves by over $\sim$ 13\%, showing that ELMo representations are indeed very strong. But adding supervised embeddings in the 1k setting further improves upon the ELMo baseline by over $\sim$ 5\%. A similar improvement of $\sim$ 5\% can be seen in the 5k setting as well. Our model shows comparable performance as the baseline when we use the entire dataset. These results suggest that the proposed supervised contextual embeddings further bring about improvements over already strong language model features in a low-resource setting. This reinforces the learning that when sufficient data is available, supervised signals do not provide information that the model cannot learn by itself from the data alone. 

\begin{table}[h]
\resizebox{\linewidth}{!}{\begin{tabular}{lccccll}
\hline
           & \multicolumn{2}{c}{\#samples=1k} & \multicolumn{2}{c}{\#samples=5k} &
        \multicolumn{2}{l}{\#samples=all}    \\ \cline{2-7} 
           & Dev             & Test           & Dev             & Test & Dev & Test            \\ \hline
Glove      & 32.30            & 33.02          &  44.98               &   46.19   & 77.62 & 77.87          \\
\begin{tabular}[c]{@{}l@{}}GloVe\\ +Ours\end{tabular}  & \textbf{37.40}            & \textbf{38.27}          & \textbf{52.11}           & \textbf{53.05}  & \textbf{77.83}  & \textbf{77.94}         \\ \hline
ELMo       & 44.69           & 45.34          &  58.30               &   58.79    & \textbf{82.68} & \textbf{82.58}      \\
\begin{tabular}[c]{@{}l@{}}ELMo\\ +Ours\end{tabular}   & \textbf{49.59}           & \textbf{50.36}          & \textbf{63.30}            & \textbf{63.84}   & 82.50 & 82.54       \\ \hline
\end{tabular}}
\caption{Performance of cross-task transfer on SRL. (\#samples=all includes $\sim$ 280K samples.) } \label{tab:srl_cross_task}
\end{table}



\paragraph{Cross-domain NER} Supervised embeddings provide an impressive 4\% improvement over the GloVe baseline with both 1,000 and 5,000 samples. Even when we replace GloVe with ELMo, we see an improvement of 3\% , indicating that the benefits of using knowledge from other domains is orthogonal to what ELMo can offer. However, the gains vanish when the full dataset is used, suggesting that knowledge from other domains is particularly useful in the very low-resource setting. However, if sufficient data is available, the model has enough resources to build upon generic word embeddings. 

Its also interesting to note that for this dataset, GloVe based models outperform their ELMo counterparts. This is probably due to the mismatch in the data used to train ELMo (formal language from the 1 billion word corpus) as opposed to the NER dataset which consists of informal language used in web blogs.

\begin{table}[h]
\resizebox{\linewidth}{!}{\begin{tabular}{lccccll}
\hline
           & \multicolumn{2}{c}{\#samples=1k} & \multicolumn{2}{c}{\#samples=5k} &
           \multicolumn{2}{l}{\#samples=all}\\ \cline{2-7} 
           & Dev             & Test           & Dev             & Test    & Dev             & Test        \\ \hline
GloVe      & 45.30           & 45.52           & 53.50           & 56.67 & 59.75 & \textbf{66.23}          \\
\begin{tabular}[c]{@{}l@{}}GloVe\\ +Ours\end{tabular} & \textbf{50.18}           & \textbf{49.64}          & \textbf{55.49}           & \textbf{60.56} & \textbf{61.16} & 65.51          \\ \hline
ELMo       & 45.06           & 45.57          & 56.43           & 57.68  & 59.58 & 64.20         \\
\begin{tabular}[c]{@{}l@{}}ELMo\\ +Ours\end{tabular}  & \textbf{48.18}           & \textbf{48.56}          &   \textbf{56.53}         & \textbf{ 57.94} & \textbf{60.36} & \textbf{65.19}          \\ \hline
\end{tabular}}
\caption{Performance of cross-domain transfer on NER (\#samples=all includes $\sim$ 17K samples)} \label{tab:ner_cross_domain}
\end{table}

\paragraph{Cross-lingual NER} In this set of experiments, we observe substantial gains by exploiting information present in other languages.
For both German and Spanish the performance gains are highest when number of samples is 1,000 , thus validating the suitability of the proposed method for transfer to very low-resource settings. Even when the entire dataset is used, we see gains over 1\% for both languages.

\begin{table}[]
\resizebox{\linewidth}{!}{
\begin{tabular}{lccccll}
\hline
                                                     & \multicolumn{2}{c}{\#samples=1k} & \multicolumn{2}{c}{\#samples=5k} & \multicolumn{2}{l}{\#samples=all} \\ \cline{2-7} 
                                                     & Dev             & Test           & Dev             & Test           & Dev             & Test            \\ \hline
  & \multicolumn{6}{c}{German} \\ \hline
MUSE                                                 & 58.12           & 57.48          & 69.85           & 67.49          & 74.53           & 71.98           \\
\begin{tabular}[c]{@{}l@{}}MUSE\\ +Ours\end{tabular} & \textbf{64.85}           & \textbf{61.60}           & \textbf{73.06}           & \textbf{70.62}          & \textbf{75.56}           & \textbf{72.96}           \\ \hline
& \multicolumn{6}{c}{Spanish} \\ \hline
MUSE                                                 & 68.05           & 71.61          & 80.48           & 82.60          & 82.01           & 82.91           \\
\begin{tabular}[c]{@{}l@{}}MUSE\\ +Ours\end{tabular} & \textbf{69.23}           & \textbf{74.59}           & \textbf{80.48}           & \textbf{82.76}          & \textbf{82.11}           & \textbf{84.22}           \\ \hline
\end{tabular}}
\caption{Performance of cross-lingual transfer on NER (\#samples=all is $\sim$ 12K for German and $\sim$ 27K for Spanish)} \label{tab:cross_ling}
\end{table}


Supervised embeddings give consistent improvements in low resource settings. Further, our approach offers an easy method to incorporate knowledge from pre-existing source models as opposed to costly training involved in other methods of knowledge transfer.
 We believe that ours is only a small step towards incorporating external supervised knowledge, and that the results are in a positive direction - paving the way for further research in aggregating  knowledge from external models. 

\section{Conclusion and Future Work}
We propose supervised contextual embeddings, an easy way to incorporate supervised knowledge from multiple pre-existing models. We perform experiments in the cross-task, cross-domain and cross-lingual setups.  

We find that ELMo embeddings provide a very strong baseline, and the proposed supervised embeddings are particularly useful in the low-resource setting. Our work points to the potential of such embeddings in various downstream tasks in different transfer learning settings. Future work includes incorporating more tasks, domains and languages, and understanding the relationships among them. These explorations would build towards our larger vision of building a more complete taxonomy of transfer learning dependencies among NLP tasks, domains and languages.

\bibliographystyle{acl_natbib}
\bibliography{acl2019}

\end{document}